\documentclass[11pt]{article}

\usepackage[preprint]{acl}

\usepackage{times}
\usepackage{latexsym}
\usepackage[T1]{fontenc}
\usepackage[utf8]{inputenc}
\usepackage{microtype}
\usepackage{inconsolata}
\usepackage{graphicx}
\usepackage{booktabs}
\usepackage{multirow}
\usepackage{amsmath}
\usepackage{amssymb}
\usepackage{enumitem}
\usepackage{xcolor}
\usepackage{adjustbox}

\newcommand{\kzero}{\mbox{$k{=}0$}}
\newcommand{\cF}{\mbox{+F}}
\newcommand{\cFI}{\mbox{+F+I}}
\newcommand{\cFIU}{\mbox{+F+I+U}}
\newcommand{\far}{FAR}
\newcommand{\oar}{OAR}
\newcommand{\repourl}{\url{https://github.com/pri2si17-1997/rad-qa-llm-eval}}

\title{Dissociating Answerability from Extraction in LLM\\ Question Answering over Radiology Reports}

\author{
  Priyanshu Sinha \\
  Labcorp, Durham, NC \\
  \texttt{sinhap3@labcorp.com} \\\And
  Kelsey McDonald \\
  Labcorp, Durham, NC \\
  \texttt{kelsey.mcdonald2@labcorp.com} \\
}

\begin{document}
\maketitle

\begin{abstract}
Answering questions from radiology reports involves two decisions: whether the report contains an answer and which span gives it.
A false answer presents an unsupported claim as fact; an over-abstention hides what the report says.
Yet RadQA results are reported as a single F1 that merges both.
We propose an evaluation method that separates them, reporting false-answer and over-abstention rates beside F1 and splitting F1 changes into extraction and answerability terms.
We apply it to 13 LLMs on the RadQA test set, zero-shot and with three cumulative demonstration sets, the last adding an unanswerable example.
Models with similar F1 differed more than twofold in false-answer rate.
The unanswerable example changed when models answered, not how well they extracted: 12 of 13 models abstained more often, and three-quarters of the answers they withdrew had F1 $\geq$ 0.5.
Within models, false answers showed no clear association with extraction quality, but acting on the decision itself helped: keeping an answer only when enough other models also answered lowered the zero-shot false-answer rate by 4 points without a detectable F1 change.
Because the two decisions can change independently, we suggest radiology QA evaluations report them separately, not through F1 alone.
\end{abstract}

\section{Introduction}

Large language models (LLMs) perform strongly on medical licensing-exam questions \citep{singhal2023large,nori2023capabilities} and medical evidence summarization \citep{tang2023evaluating}, and are increasingly applied to patient records, from answering clinicians' questions about them \citep{fleming2024medalign} to extracting information from radiology reports \citep{leguellec2024performance}.
Radiology reports are a natural target for LLM question answering: they are produced in large volumes, use dense domain terminology, and are read by ordering physicians who need accurate answers from them.
Exam benchmarks test medical knowledge, not the reading of a specific report, where the answer must be found in a section such as Findings or Impression and is sometimes not there at all.

Extractive question answering (QA) over reports is a controlled version of this setting: the answer is a verbatim span of the report, or the model declines because it decides the question is unanswerable.
In RadQA \citep{soni-etal-2022-radqa}, physicians wrote questions from the clinical referral of MIMIC-III radiology reports \citep{PhysioNet-mimiciii-1.4,johnson2016mimic} without seeing the report text, so a quarter of its test questions have no answer in the report. Such a system makes two decisions per question: whether the report answers it, and if so, which span does.
The two fail in clinically distinct ways.
A \emph{false answer} to an unanswerable question puts a statement the report does not support in front of a clinician; an \emph{over-abstention} withholds information the report contains and sends the clinician back to the full text.

The standard metric, SQuAD 2.0 F1 \citep{rajpurkar-etal-2018-know}, folds both decisions into one number: a correct abstention scores 1, a wrong answer/abstain decision scores 0, and otherwise the score is token overlap between words in a predicted answer and the ground-truth answer. Models with the same F1 can therefore give very different numbers of false answers, and a change in F1 between prompts can come from better spans or from a moved abstention boundary.
SQuAD 2.0's evaluation does also report F1 separately on answerable and unanswerable questions (HasAns/NoAns), and NoAns F1 is exactly one minus the false-answer rate.
But RadQA results are reported only as overall F1 and EM \citep{soni-etal-2022-radqa,al2025radqa}.
Even HasAns F1 mixes the two decisions: it equals the share of answerable questions answered times the mean F1 of those answers, so it changes when either one does.

The distinction matters most for in-context demonstrations, the obvious intervention that needs no training.
On RadQA, LLMs still trail task-specific fine-tuning: RadQA-DPO \citep{al2025radqa} reaches 77.48 test F1 and 55.15 exact match (EM), and the two annotators reach 78.72 and 90.31 F1 \citep{soni-etal-2022-radqa}.
In-context demonstrations \citep{brown2020language} could narrow this gap if they improve extraction, but not if they mainly move the answer/abstain boundary.
Studies of demonstration choice, order and format report effects on aggregate accuracy \citep{zhao2021calibrate,lu-etal-2022-fantastically,min-etal-2022-rethinking}, which cannot tell which decision moved.

\begin{figure}[t]
  \centering
  \includegraphics[width=\columnwidth]{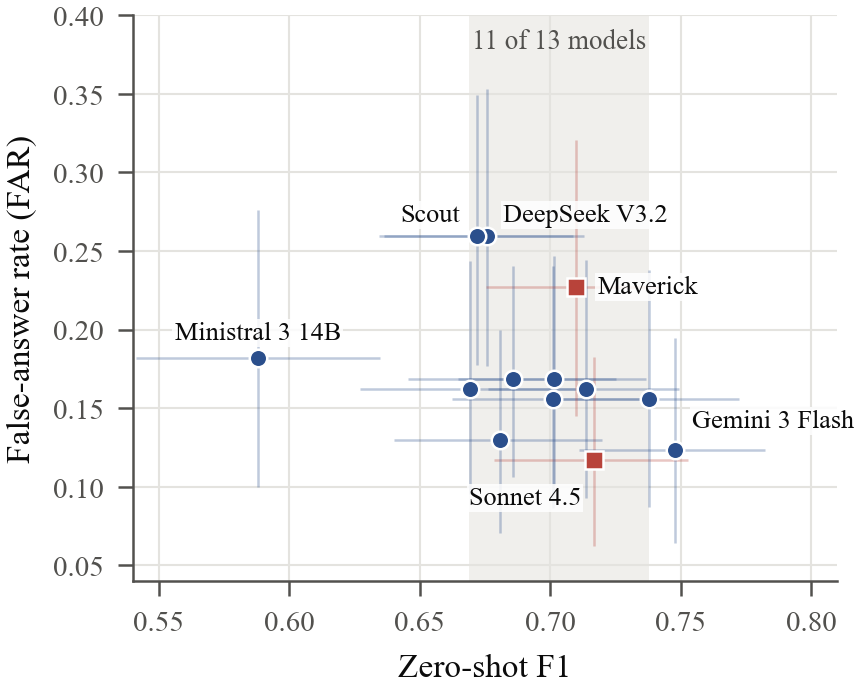}
  \caption{Zero-shot span F1 against false-answer rate (\far{}) for the 13 LLMs, with report-clustered 95\% CIs. The shaded band holds 11 models within 0.07 F1 that differ more than twofold in \far{}. Squares mark the example pair discussed in \S\ref{sec:res-zeroshot}: Claude Sonnet~4.5, the lowest-\far{} model, and LLaMA~4 Maverick, the highest-F1 model whose \far{} is significantly higher than Sonnet~4.5's. They differ by 0.007 in F1 but 0.110 in \far{}. Values for unlabelled models are in Table~\ref{tab:zeroshot}.}
  \label{fig:f1-far}
\end{figure}

We propose an evaluation method that separates the two decisions, reporting the false-answer rate (\far{}) and over-abstention rate (\oar{}) beside span F1, and apply it to 13 LLMs from six families on the RadQA test set, zero-shot and with three cumulative demonstration sets (an answerable Findings question, an answerable Impression question, and an unanswerable question).
Our contributions:

\begin{enumerate}[leftmargin=*,itemsep=1pt,topsep=2pt]
  \item \textbf{A decomposition of $\Delta$F1 that isolates extraction, with a selectivity test} (\S\ref{sec:scoring}). Any change in F1 splits into an extraction term, computed on answerable questions answered in both conditions, and an answerability shift. A companion test asks whether new abstentions remove weak answers or good ones. Unlike a change in HasAns F1, the extraction term contains no abstentions, and both tools work with any span metric. \far{} and \oar{} follow from the standard HasAns/NoAns breakdown, so we offer them as a reporting recommendation rather than as new metrics.
  \item \textbf{The tested unanswerable demonstration moves the answer decision, not extraction} (\S\ref{sec:res-demos}). It raises \oar{} in 12 of 13 models. Of the resulting $-0.0126$ change in F1, $-0.0127$ is answerability shift, and extraction is flat under both token and RadGraph F1 \citep{jain2021radgraph}. For answerable demonstrations, the split depends on the span metric.
  \item \textbf{The added abstentions are not detectably selective} (\S\ref{sec:res-selectivity}). Three-quarters of the 272 withdrawn answers had F1 $\geq$ 0.5. In exchange, models gave a net 55 fewer false answers, most of them from one model.
  \item \textbf{Similar F1, different false-answer rates} (\S\ref{sec:res-zeroshot}, \S\ref{sec:res-far-accuracy}; Figure~\ref{fig:f1-far}). Eleven of 13 models lie within 0.669--0.738 zero-shot F1, while \far{} ranges from 0.117 to 0.260. The cross-model F1--\far{} correlation is largely built into the metric, and within a model we find no association between extraction quality and false answers.
  \item \textbf{Evaluating a mitigation that targets the answer decision} (\S\ref{sec:res-gate}). A consensus gate, cross-fitted over reports, lowers zero-shot \far{} by 4 points with no detectable F1 change.
\end{enumerate}

\noindent We also find that no model matches the exact match of the fine-tuned RadQA-DPO, and none reaches inter-annotator agreement (\S\ref{sec:res-zeroshot}).
Our code is available at \repourl{}.

\section{Related Work}

\paragraph{Clinical and radiology QA.}
Domain-adapted encoders such as BioBERT, ClinicalBERT and RadBERT advanced extraction and QA over clinical text \citep{Lee_2019_biobert,alsentzer2019publicly,huang2020clinicalbertmodelingclinicalnotes,yan2022radbert}, and radiology-specific models followed \citep{ghosh-etal-2023-radling,ranjit2024radphi3smalllanguagemodels}.
emrQA \citep{pampari-etal-2018-emrqa} and RadQA \citep{soni-etal-2022-radqa} provide extractive QA over clinical notes and radiology reports.
In RadQA, questions were written from the clinical referral without seeing the report, so unanswerable questions arise naturally rather than by construction.
Fine-tuned RadQA systems have progressed from BERT baselines \citep{soni-etal-2022-radqa} to RadQA-DPO \citep{al2025radqa}.
General and medical LLMs perform strongly on exam-style medical QA \citep{singhal2023large,singhal2025toward,nori2023capabilities,chen2023meditron,labrak2024biomistral,xie2024me}, and have been applied to clinical extraction and radiology reporting \citep{busch2025large,agrawal-etal-2022-large}.
Like prior RadQA work, these evaluations report span F1 and EM; we separate the answer decision from extraction and ask which of the two demonstrations change.

\paragraph{Unanswerable questions and abstention.}
SQuAD 2.0 \citep{rajpurkar-etal-2018-know} added unanswerable questions, and its official evaluation reports F1 separately on answerable and unanswerable questions (HasAns/NoAns).
Selective QA lets a model abstain when it is likely to be wrong and evaluates the resulting trade-off between coverage and accuracy \citep{kamath-etal-2020-selective}; related work asks when LLMs decline to answer and whether they know what they know \citep{wen2025know,kadavath2022language}.
This breakdown already exposes false answers, but HasAns F1 scores an over-abstention as 0 and so mixes span quality with abstention; our extraction term excludes abstentions.
We also ask the selective-prediction question of a prompting intervention: do the abstentions it adds fall on weak answers?

\paragraph{Sensitivity of in-context learning.}
In-context learning \citep{brown2020language} is sensitive to the choice, order and label balance of demonstrations \citep{zhao2021calibrate,lu-etal-2022-fantastically}.
\citet{min-etal-2022-rethinking} find that demonstrations act largely through the label space, input distribution and format they expose, rather than through correct input--label mappings.
Our results show a related pattern for extractive clinical QA: the unanswerable demonstration changes how often models use the abstain option, not how well they extract.

\paragraph{Evaluation reliability.}
Comparisons of NLP systems benefit from significance testing that respects the structure of the test data \citep{dror-etal-2018-hitchhikers}.
For clinical content, RadGraph \citep{jain2021radgraph} annotates entities and relations in radiology reports \citep{yu2023evaluating}.
We use RadGraph F1 as an entity-level complement to token F1 and report-clustered inference throughout.

\section{Methods}

\subsection{Data}
\label{sec:data}

We use the RadQA test split \citep{soni-etal-2022-radqa,soni2022radqaphysionet}: 614 questions over 104 MIMIC-III radiology reports.
Each answer is either a text span within the Findings or Impression section, or the question is marked unanswerable. 

Of the 614 questions, 154 (25.1\%) are unanswerable, across 57 reports. The remaining 460 answerable questions are drawn from the Findings section (255 questions) and the Impression section (205 questions). The distributed file has one row per question and annotator answer (894 rows), and 280 answerable questions carry two annotators' answers.
We score one prediction per question (the first of the two rows for doubly annotated questions) and analyze at the question level (614 questions); only token usage (Table~\ref{tab:tokens}) is reported per row (894 rows).

\subsection{Models and Inference}

We evaluate 13 models from six families (Table~\ref{tab:models}): Claude Opus~5, Sonnet~5, Sonnet~4.5 and Haiku~4.5; Gemini~3.1 Pro and 3 Flash; LLaMA~4 Maverick and Scout; DeepSeek V3.2; Mistral Large~3 and Ministral~3 14B; and GPT-OSS 120B and 20B, through Google Vertex AI (Gemini) and AWS Bedrock (all others).

Each family has its own system prompt, and all share one task specification: return the exact, minimal verbatim span that answers the question, or mark it unanswerable.
The output is a JSON object with the extracted spans, a Boolean answerability flag and a short rationale, enforced through tool calling (LLaMA~4 output is parsed from free text).
Calls that still fail after three attempts are scored as abstentions (3 of 31,928 predictions), so failures cannot raise \far{}.

All models ran with a 4,096-token output limit. All except Claude Opus~5 and Sonnet~5 (11 of 13) ran at temperature 0; those two ran at the provider-default temperature, as user-defined temperature parameters were deprecated. Each model used its provider's default reasoning and thinking settings. 

\subsection{Demonstration Conditions}
\label{sec:conditions}

Demonstrations are drawn from the RadQA train split (reports of at most 4,000 characters) and are fixed across all demonstration conditions. We select one example of each type: an answerable Findings question, an answerable Impression question and an unanswerable question, each paired with a fixed rationale. 

The four conditions are cumulative: \kzero{} (none); \cF{} (an answerable Findings question); \cFI{} (adding an answerable Impression question); and \cFIU{} (adding an unanswerable question, placed last).
The unanswerable demonstration is a near-miss case: it asks about gallbladder stones in a report describing a common bile duct stone after cholecystectomy, and its rationale states that the information is not mentioned.
Every model and test question receives the same demonstrations. Because number and type change together, we attribute effects to the demonstration set, not solely to its size.

\subsection{Scoring and Decomposition}
\label{sec:scoring}

Following SQuAD 2.0 \citep{rajpurkar-etal-2018-know}, a question scores 0 if prediction and ground truth disagree on answerability, 1 if both say unanswerable, and otherwise the maximum token F1 (or EM) over ground truth answers and predicted spans.

A prediction's answerability is its \texttt{answerable} flag, so a prediction flagged answerable with no span counts as an answer: it scores 0 on an answerable question and is a false answer on an unanswerable one.

\emph{RadGraph F1} \citep{jain2021radgraph} measures the overlap of clinical entities and relations between the predicted and reference answer spans, with answerability treated the same way and the maximum taken over ground truth answers. Unlike token F1, it scores a multi-span prediction as one string, with the spans joined, rather than crediting the best-matching span; about 6\% of answers to answerable questions contain more than one span. It is among the automatic metrics that agree best with radiologists' error counts on generated reports \citep{yu2023evaluating}.
We use it as an entity-level complement to token F1 that is less sensitive to span boundaries, and report each decomposition under both metrics to check whether conclusions hold for clinical content as well as for token overlap.

\paragraph{Answer-decision metrics.}
The \emph{false-answer rate} (\far{}) is the share of the 154 unanswerable questions a model answers; the \emph{over-abstention rate} (\oar{}) is the share of the 460 answerable questions it declines.
Both are measured against the physician annotators' answerability labels and describe the answer/abstain decision; they are not measures of clinical harm.

\paragraph{Decomposition of $\Delta$F1.}
For one model and conditions $A \rightarrow B$, let $s_q^A$ and $s_q^B$ be the scores of question $q$, and let $\mathcal{E}$ be the answerable questions the model answers in both conditions.
With $d_q = s_q^B - s_q^A$ and $n = 614$ questions,
\begin{equation}
  \Delta = \underbrace{\tfrac{1}{n}\textstyle\sum_{q \in \mathcal{E}} d_q}_{\text{extraction}}
  \;+\; \underbrace{\tfrac{1}{n}\textstyle\sum_{q \notin \mathcal{E}} d_q}_{\text{answerability shift}} .
  \label{eq:decomp}
\end{equation}
The extraction term captures span-level changes for questions where the model's answer decision stays the same. The answerability shift term captures all cases where the decision itself changes, such as a new abstention, a newly answered question, or a flipped decision on an unanswerable question.

\paragraph{Selectivity of new abstentions.}
Let $\mathcal{N}$ be the answerable questions answered in $A$ but declined in $B$; each contributes $-s_q^A/n$ to the answerability term.
We compare the mean prior score $s_q^A$ over $\mathcal{N}$ with that over $\mathcal{E}$.
If newly abstained questions had lower prior scores, the abstentions are \emph{selective}: they remove weak answers; if prior scores were higher, then they removed valid answers.

\paragraph{Within-model predictors of false answers.}
To ask whether a model's false answers follow from weak extraction, we use the 52 reports that contain both answerable and unanswerable questions (126 unanswerable questions).
For each model and report, \emph{extraction quality} is the model's mean F1 on the report's answerable questions it answered, and \emph{answering propensity} is the fraction of them it answered.
We relate both to whether the model answers each of the report's unanswerable questions, controlling for report difficulty (the other 12 models' mean F1 on the report; Appendix~\ref{app:stats}).

\paragraph{Other measures.}
Because SQuAD F1 credits correct abstentions, $\text{F1} = \bigl(\sum_{q\,\text{answerable}} s_q + 154\,(1-\far{})\bigr)/614$, so its correlation with \far{} is partly built in.
We therefore also use \emph{answerable F1} (mean over the 460 answerable questions, abstentions scored 0) and \emph{extraction F1} (mean over the answerable questions the model answered).
In SQuAD 2.0's breakdown, NoAns F1 $= 1-\far{}$ and HasAns F1 (our answerable F1) $= (1-\oar{}) \times$ extraction F1.
An error taxonomy assigns each question to exact match, EM-mismatch (F1 $\geq$ 0.5, EM $=$ 0), partial ($0.1 \leq$ F1 $< 0.5$), no overlap (F1 $< 0.1$), correct abstention, over-abstention or false answer.
Published human, BERT-MIMIC and RadQA-DPO results \citep{soni-etal-2022-radqa,al2025radqa} report token F1 and EM only, so comparisons with them are descriptive.
For a like-for-like human reference under all three metrics, we score models and annotators against single annotators on the 280 doubly annotated questions.

\subsection{Consensus Gate}
\label{sec:gate}

To test whether the answer decision can be improved without changing extraction, we gate each model's answers with the other models' decisions.
For a model in a given condition, let $v_q$ be the number of the other 12 models, in the same condition, that answer question $q$.
The gate keeps the model's span if $v_q \geq t$ and otherwise replaces it with an abstention, scored 1 if $q$ is unanswerable and 0 otherwise.
Abstentions are never overridden and kept spans are unchanged, so the gate can only lower \far{} and raise \oar{}.
At $t = 0$ the model is ungated; at $t = 12$ an answer is kept only if all 12 other models also answer.
Because we have predictions only for the test split, $t$ is cross-fitted: we split the 104 reports at random into halves, choose $t$ on one half by maximizing span F1, and measure the change from the ungated model on the other.
$t$ is chosen per model (own-$t$) or as one value for all models (pooled-$t$).
We repeat this over 200 splits and report the mean model-mean change, with the 2.5--97.5\% range over splits.
A two-model gate instead keeps an answer only if one partner model also answers; we report the median over the 156 ordered pairs.

\subsection{Statistical Analysis}
\label{sec:stats}

Inference is clustered by report (104 clusters).
Confidence intervals are report-level cluster-bootstrap intervals (10,000 replicates; 2,000 for decomposition terms).
Per-model condition effects use Wilcoxon signed-rank tests on paired report means, with Benjamini--Hochberg (BH) correction \citep{benjamini1995controlling} across models.
For \far{}, we use Cochran's Q across models and, against the lowest-\far{} model, exact McNemar and report-level permutation tests.
Selectivity and within-model predictors of false answers use generalized estimating equations (GEE) \citep{liang1986longitudinal} with exchangeable within-report correlation.
Across the 13 models, we report Spearman $\rho$ with bootstrap CIs and leave-one-model-out checks, and interpret these descriptively.
Full specifications are in Appendix~\ref{app:stats}.

\section{Results}

\begin{table*}[t]
  \centering\small
  \setlength{\tabcolsep}{5pt}
  \begin{adjustbox}{max width=\linewidth}
  \begin{tabular}{lccccc}
    \toprule
    Model & F1 [95\% CI] & EM & RadGraph F1 & \far{} [95\% CI] & \oar{} \\
    \midrule
    Gemini 3 Flash      & 0.748 [0.711, 0.782] & 0.482 & 0.673 & 0.123 [0.064, 0.195] & 0.100 \\
    Gemini 3.1 Pro      & 0.738 [0.702, 0.772] & 0.451 & 0.654 & 0.156 [0.087, 0.238] & 0.083 \\
    Claude Sonnet 4.5   & 0.717 [0.678, 0.753] & 0.436 & 0.659 & 0.117 [0.062, 0.182] & 0.128 \\
    Claude Opus 5   & 0.714 [0.676, 0.749] & 0.386 & 0.628 & 0.162 [0.093, 0.245] & 0.057 \\
    LLaMA 4 Maverick    & 0.710 [0.675, 0.742] & 0.381 & 0.620 & 0.227 [0.145, 0.321] & 0.061 \\
    Claude Sonnet 5 & 0.701 [0.665, 0.737] & 0.388 & 0.623 & 0.169 [0.101, 0.247] & 0.096 \\
    Claude Haiku 4.5    & 0.701 [0.662, 0.738] & 0.409 & 0.639 & 0.156 [0.086, 0.240] & 0.135 \\
    Mistral Large 3     & 0.686 [0.645, 0.725] & 0.410 & 0.614 & 0.169 [0.107, 0.240] & 0.133 \\
    GPT-OSS 120B        & 0.681 [0.640, 0.720] & 0.383 & 0.602 & 0.130 [0.071, 0.200] & 0.150 \\
    DeepSeek V3.2       & 0.676 [0.636, 0.713] & 0.389 & 0.618 & 0.260 [0.177, 0.353] & 0.130 \\
    LLaMA 4 Scout       & 0.672 [0.634, 0.709] & 0.375 & 0.585 & 0.260 [0.178, 0.349] & 0.100 \\
    GPT-OSS 20B         & 0.669 [0.627, 0.709] & 0.376 & 0.585 & 0.162 [0.096, 0.244] & 0.141 \\
    Ministral 3 14B     & 0.588 [0.541, 0.635] & 0.358 & 0.535 & 0.182 [0.100, 0.276] & 0.202 \\
    \midrule
    \multicolumn{6}{l}{\emph{Published references (RadQA test set; not re-run)}} \\
    Human, best annotator$^a$       & 0.903 & 0.814 & -- & -- & -- \\
    Human, mean of two annotators$^a$ & 0.845 & 0.754 & -- & -- & -- \\
    RadQA-DPO, Flan-T5-3B$^b$       & 0.775 & 0.552 & -- & -- & -- \\
    BERT-MIMIC$^a$                  & 0.636 & 0.494 & -- & -- & -- \\
    \bottomrule
  \end{tabular}
  \end{adjustbox}
  \caption{Zero-shot (\kzero{}) results on the RadQA test set (614 questions). CIs are report-clustered bootstrap intervals (10,000 replicates). \far{}: share of the 154 unanswerable questions answered; \oar{}: share of the 460 answerable questions declined. References: $^a$\citet{soni-etal-2022-radqa}; $^b$\citet{al2025radqa}. Published references use their authors' scoring and are not directly comparable (\S\ref{sec:scoring}).}
  \label{tab:zeroshot}
\end{table*}

\subsection{Zero-Shot Performance}
\label{sec:res-zeroshot}

\paragraph{Similar F1, different false-answer rates.}
Zero-shot F1 ranges from 0.588 to 0.748, and 11 of 13 models lie between 0.669 and 0.738, with overlapping CIs for the top ten (Table~\ref{tab:zeroshot}).
\far{} ranges from 0.117 to 0.260, and the models differ overall (Cochran's $Q = 69.2$, df $= 12$, $p < 10^{-5}$).
Against the lowest-\far{} model, Claude Sonnet~4.5 (0.117), three models are significantly higher under both the McNemar and the cluster permutation test: LLaMA~4 Maverick (0.227; $q = 0.0001$ and $0.004$), DeepSeek V3.2 (0.260; $q = 0.0002$ and $0.011$) and LLaMA~4 Scout (0.260; $q < 0.0001$); all others have $q \geq 0.06$.
Maverick thus has the fifth-highest F1 but one of the three highest \far{}s (Figure~\ref{fig:f1-far}).
\oar{} ranges from 0.057 to 0.202.

\paragraph{Where F1 is lost.}
Pooled over models at \kzero{}, 19.5\% of questions are exact matches, 35.4\% EM-mismatches, 9.4\% partial, 1.9\% no overlap, 20.7\% correct abstentions, 8.7\% over-abstentions and 4.4\% false answers.
84.9\% of EM-mismatches share clinical entities with the correct answer (RadGraph F1 $> 0$), whereas only 2.1\% of the 95 non-empty no-overlap answers do.
Answers that miss the correct answer entirely are therefore rare.
Most lost F1 comes from span boundaries and from the answer decision.

\paragraph{Gap to fine-tuning and to humans.}
No model reaches RadQA-DPO's F1 (0.775) in any condition; only Gemini~3 Flash at \kzero{} has a CI that includes it, and all 52 model $\times$ condition EM CIs lie below its 0.552.
On the 280 doubly annotated questions, every model is below inter-annotator agreement in every condition under all three metrics (BH $q < 0.0001$).
At \kzero{}, models score a mean token F1 of 0.756, EM of 0.318 and RadGraph F1 of 0.625 against single annotators; the annotators score 0.935, 0.746 and 0.858 against each other.
Models thus reach 81\% of human agreement under token F1 but only 73\% under RadGraph F1, so the shortfall is not only a matter of span boundaries.

\subsection{What Demonstrations Change}
\label{sec:res-demos}

\begin{figure}[t]
  \centering
  \includegraphics[width=\columnwidth]{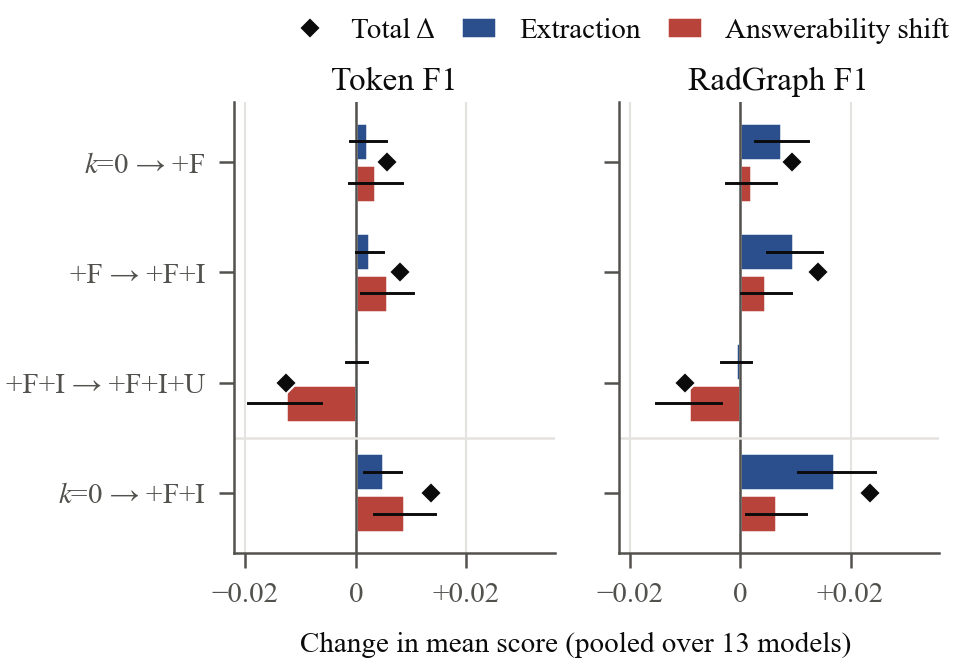}
  \caption{Decomposition of the change in mean score between conditions (Eq.~\ref{eq:decomp}), pooled over 13 models, with 95\% report-cluster bootstrap CIs. Bars sum to the total, shown as $\blacklozenge$. Exact values are in Table~\ref{tab:decomp}.}
  \label{fig:decomp}
\end{figure}

\paragraph{Small changes in F1.}
Model-mean F1 is 0.692 at \kzero{}, 0.698 at \cF{}, 0.706 at \cFI{} and 0.693 at \cFIU{} (per-model values in Table~\ref{tab:conditions}).
Answerable demonstrations raise F1 in 10 of 13 models (Wilcoxon over model means, $p = 0.048$), but only DeepSeek V3.2 and Ministral~3 14B improve significantly under report-clustered testing.
Adding the unanswerable demonstration lowers F1 in 12 of 13 models ($p = 0.0034$; significant for four models), raises mean \oar{} from 0.105 to 0.142 and lowers mean \far{} from 0.171 to 0.143.

\paragraph{The unanswerable demonstration moves the answer decision.}
Of the $-0.0126$ token-F1 drop from \cFI{} to \cFIU{}, $-0.0127$ is answerability shift and $+0.0001$ extraction (Figure~\ref{fig:decomp}).
Extraction does not change significantly in any of the 13 models (BH $q \geq 0.48$), while the answerability shift is negative in 12 of the 13.
On answerable questions, the demonstration adds 272 abstentions.
On unanswerable questions, it adds 18 false answers and removes 73, a net reduction of 55.
Under RadGraph F1 the picture is the same: $-0.0093$ of the $-0.0101$ drop is answerability shift, and extraction is flat ($-0.0008$ [$-0.0034$, $+0.0019$]; BH $q \geq 0.40$ in every model).

\paragraph{For answerable demonstrations, the split depends on the metric.}
Under token F1, 64\% of the $+0.0136$ gain from \kzero{} to \cFI{} is answerability shift, and per model, extraction improves significantly only for Ministral~3 14B.
Under RadGraph F1, 73\% of the $+0.0235$ gain is extraction, positive in all 13 models and significant in 6.
The two metrics rank models almost identically ($\rho$ 0.87--0.93) and disagree only on spans answered in both conditions ($+0.0075$ token vs.\ $+0.0265$ RadGraph F1).
The span convention explains little of this: scoring token F1 on the joined prediction, as RadGraph F1 does (\S\ref{sec:scoring}), raises its extraction term only to $+0.0063$ [$+0.0030$, $+0.0096$], or 41\% of its gain.
Without radiologist adjudication we cannot say which better reflects clinical correctness, so we do not claim that answerable demonstrations act mainly through answerability.

\subsection{Which Answers Do New Abstentions Remove?}
\label{sec:res-selectivity}

\begin{table}[t]
  \centering\small
  \setlength{\tabcolsep}{3pt}
  \begin{adjustbox}{max width=\linewidth}
  \begin{tabular}{lcccccc}
    \toprule
    & & \multicolumn{2}{c}{Prior F1} & & & \\
    \cmidrule(lr){3-4}
    Transition & $|\mathcal{N}|$ & $\mathcal{N}$ & $\mathcal{E}$ & Diff. & $p$ & $\geq$0.5 \\
    \midrule
    \kzero{}$\rightarrow$\cF{} & 134 & 0.618 & 0.736 & $-$0.099 & 0.0065 & 67\% \\
     & & & & {\scriptsize [$-$0.170, $-$0.028]} & & \\
    \cF{}$\rightarrow$\cFI{} & 96 & 0.676 & 0.738 & $-$0.042 & 0.30 & 68\% \\
     & & & & {\scriptsize [$-$0.123, $+$0.038]} & & \\
    \cFI{}$\rightarrow$\cFIU{} & 272 & 0.701 & 0.745 & $-$0.035 & 0.16 & 76\% \\
     & & & & {\scriptsize [$-$0.084, $+$0.014]} & & \\
    \midrule
    \kzero{}$\rightarrow$\cFI{} & 144 & 0.604 & 0.737 & $-$0.121 & 0.0014 & 61\% \\
     & & & & {\scriptsize [$-$0.196, $-$0.047]} & & \\
    \bottomrule
  \end{tabular}
  \end{adjustbox}
  \caption{Selectivity of new abstentions on answerable questions, pooled over 13 models. $\mathcal{N}$: questions newly declined; $\mathcal{E}$: questions answered in both conditions. Prior F1 is F1 in the earlier condition. Diff.\ (95\% CI below) is from a GEE of prior F1 on a newly-declined indicator with model fixed effects and report clustering. $\geq$0.5: share of $\mathcal{N}$ with prior F1 $\geq$ 0.5.}
  \label{tab:selectivity}
\end{table}

\paragraph{The unanswerable demonstration is not detectably selective.}
Its 272 new abstentions had a mean prior F1 of 0.701, against 0.745 for questions still answered (Table~\ref{tab:selectivity}).
The CI of the difference includes zero, although it cannot exclude modest selectivity.
This holds when any single model is left out (leave-one-out $p \geq 0.08$), under RadGraph F1 ($-0.047$ [$-0.107$, $+0.013$]) and in every model separately (all $q > 0.24$).
Three-quarters had F1 $\geq$ 0.5, so its F1 cost comes mostly from withdrawing good answers.

\paragraph{Answerable-demonstration selectivity comes from one model.}
From \kzero{} to \cFI{}, newly declined questions had a prior F1 0.12 lower than those still answered, but this rests on Ministral~3 14B (22 new abstentions with prior F1 0.083 vs.\ 0.676 kept; $q < 0.0001$).
Without it, the pooled difference is $-0.042$ ($p = 0.27$), and LLaMA~4 Maverick goes the other way ($+0.146$, $q = 0.017$).
RadGraph F1 agrees on the pooled result.
We therefore treat this selectivity as model-specific.

\subsection{How False Answers Relate to Accuracy}
\label{sec:res-far-accuracy}

\paragraph{Across models} (Table~\ref{tab:corr}).
At \kzero{}, F1 and \far{} correlate ($\rho = -0.59$ [$-0.86$, $-0.06$]), but largely because F1 credits correct abstentions.
The \far{} range alone accounts for 0.036 of the 0.160 F1 range.
For 7 of the 13 models, leaving that one model out makes the correlation non-significant, and with answerable F1 in place of F1 it is only $-0.28$ ($p = 0.35$).
Extraction F1 and \far{} are only weakly associated ($\rho$ $-0.43$ to $-0.59$, significant only at \cF{}).
With demonstrations, models with lower \far{} instead have higher \oar{} ($\rho$ $-0.58$ to $-0.72$, against $-0.08$ at \kzero{}; robust to leaving out any model at \cFI{} and \cFIU{}).
This is consistent with models differing mainly in how readily they answer.
With demonstrations, no F1--\far{} correlation is distinguishable from zero ($\rho$ $-0.23$ to $+0.25$), so F1 rank is not a dependable guide to false answers.

\paragraph{Within models} (Table~\ref{tab:gee}).
On reports with both answerable and unanswerable questions, a model's extraction quality on the answerable ones is not associated with a false answer on the unanswerable ones in any condition (OR per SD 0.94--1.04; upper CI bounds $\leq 1.18$). With about four answerable questions per report, this measure is noisy, which biases the association toward null, so we cannot exclude a modest effect.
A model's answering propensity on the report is nominally associated with false answers (OR 1.21 at \kzero{}, 1.31 at \cFI{}; BH $q = 0.076$), which we treat as suggestive.

\subsection{Consensus Gating of the Answer Decision}
\label{sec:res-gate}

With $t$ chosen on held-in reports (\S\ref{sec:gate}), at \kzero{} the gate lowers mean \far{} by 0.040 [0.026, 0.057] on held-out reports and raises \oar{} by 0.010, with no detectable F1 change ($+0.005$ [$-0.000$, $+0.009$]; Table~\ref{tab:gate}).
The effect is concentrated in high-\far{} models (DeepSeek V3.2 $-0.144$, LLaMA~4 Scout $-0.099$, Ministral~3 14B $-0.076$, LLaMA~4 Maverick $-0.063$).
It is smallest at \cFIU{} ($-0.011$), where the unanswerable demonstration has already moved models toward abstention.
Two-model gates do worse: the median of the 156 ordered pairs lowers \far{} by 0.049 but costs 0.013 F1.

\section{Discussion}

\paragraph{Two errors with different consequences, and F1 separates neither.}
A false answer puts a statement the report does not support in front of the reader; an over-abstention sends the reader back to the full report.
How much each costs depends on the use.
A system that pre-fills structured fields for later review \citep{adams2023leveraging} and one that answers a clinician's question about a specific patient \citep{fleming2024medalign} weigh them differently, and neither weighting is a property of the benchmark.
When models are compared or selected for clinical QA, \far{} and \oar{} at the operating point carry information that F1 rank does not.

\paragraph{The unanswerable demonstration moved the answer threshold.}
The largest and most consistent demonstration effect was on the answer decision.
Pooled over models, it withdrew about three good answers (F1 $\geq$ 0.5) for each false answer it removed (\S\ref{sec:res-selectivity}).
The reduction is uneven: LLaMA~4 Scout accounts for 34 of the net 55 fewer false answers, and \far{} fell in only 7 of the 13 models while \oar{} rose in 12 of the 13.
Whether that exchange is acceptable depends on the relative cost of the two errors, which only the clinical setting can fix.
Prompt construction thus sets an operating point implicitly.
For clinical QA, we suggest setting it explicitly against a stated \far{} target and re-checking it with the decomposition and selectivity test whenever the prompt changes.
With one demonstration set, this is evidence about this demonstration only.
Testing several unanswerable demonstrations, from clean absences to near misses, each with its own rationale and with count-matched answerable controls, would show whether the threshold shift generalizes.

\paragraph{Better extraction may not reduce false answers.}
Within a model, we found no association between extraction quality on a report and whether the model answers that report's unanswerable questions, although the per-report measure is noisy and modest effects cannot be excluded.
This suggests that improving extraction alone is unlikely to reduce false answers much; as a within-model result, it does not by itself show how models with different F1 differ in false answers.
The answer decision may therefore need its own control, such as a verification or consensus step or a separately calibrated answerability model \citep{kamath-etal-2020-selective,wen2025know}.
Our gate shows that information from outside the model can lower \far{} without a detectable F1 change, mostly in the highest-\far{} models.
It shows that the answer decision can be improved independently of extraction; it is not a deployment recipe.
In practice, such a gate would need a calibration set separate from the evaluation data, a \far{} target chosen by the system's clinical owners, recalibration on local data, and 13 inference calls per question.

\paragraph{Where the remaining gap lies.}
Demonstrations changed model-mean F1 by at most 0.014, while the gap to RadQA-DPO and to the annotators lies in the spans themselves: exact boundaries and, under RadGraph F1, the clinical entities extracted.
We offer this as a hypothesis about what span-level supervision supplies and a few demonstrations do not.
Where the exact extracted text matters, the models evaluated here do not yet substitute for a fine-tuned extractor or human review.

\paragraph{Reporting recommendations.}
For extractive clinical QA with unanswerable questions, we recommend reporting
(i)~\far{} and \oar{} beside span F1;
(ii)~an F1 restricted to answerable questions;
(iii)~the decomposition of Eq.~\ref{eq:decomp}, with the selectivity of new abstentions, when comparing prompts or models;
and (iv)~at least two span metrics, since token and RadGraph F1 disagree on whether answerable demonstrations improve extraction, a disagreement that comes mostly from what the metrics measure rather than from their span conventions (\S\ref{sec:res-demos}).
This matters when a prompt or model change is approved for deployment: under token F1 alone the answerable demonstrations barely improve extraction, whereas RadGraph F1 shows a gain in all 13 models, significant in 6.
The approving metric should match how the output is used (entity-level agreement when findings populate structured fields, token F1 and EM when the extracted text is shown to clinicians), and since neither has been checked against radiologists on short spans, local clinical review is still needed.

\section{Conclusion}

The overall F1 reported for RadQA folds two decisions into one number; we separated them with an extraction term restricted to unchanged answer decisions and a test of which answers new abstentions remove.
On RadQA, the unanswerable demonstration we tested moved the answer decision rather than extraction: it raised over-abstention in 12 of 13 LLMs, and three-quarters of the answers it withdrew had F1 $\geq$ 0.5.
At similar zero-shot F1, models differed more than twofold in false-answer rate, and within a model, we found no association between extraction quality and false answers.
Evaluations of clinical QA systems should therefore report \far{} and \oar{} beside span F1, and check any prompt or model change with the decomposition and selectivity test before deployment is considered.

\section*{Limitations}

\paragraph{No clinical adjudication.}
\far{} and \oar{} measure agreement with the physician annotators' answerability labels, not whether an answer would mislead a clinician.
Token and RadGraph F1 are automatic measures of span agreement, and RadGraph F1 was validated against radiologists on full reports, not short spans.
The labels themselves are imperfect \citep{soni-etal-2022-radqa}, so some questions counted as unanswerable may be answerable, or the reverse.
No radiologist reviewed correctness or clinical consequence.
A stratified adjudication of EM mismatches, token/RadGraph disagreements, false answers and newly induced abstentions is the missing validation.

\paragraph{One dataset and one demonstration set.}
All results come from the RadQA test split: one institution, one report style, and 154 unanswerable questions in 57 reports.
All few-shot results use one fixed demonstration set with a single, near-miss unanswerable example whose generic rationale does not describe the near miss (\S\ref{sec:conditions}).
Because each condition adds a new demonstration type, we cannot separate ``a third demonstration'' from ``an unanswerable demonstration'', or this example from unanswerable examples in general.

\paragraph{Thirteen models.}
Cross-model correlations rest on 13 points, and several are sensitive to single models (\S\ref{sec:res-far-accuracy}).

\paragraph{Consensus gate.}
The gate threshold is cross-fitted within the test set (\S\ref{sec:gate}), not set on a separate calibration split, and the gate is built from and evaluated on the same 13 models.

\paragraph{External comparisons.}
Human, BERT-MIMIC and RadQA-DPO results are taken from their papers.
Their test answers, like ours, keep both annotators' answers where both answer the question \citep{soni-etal-2022-radqa}, and RadQA-DPO is scored with the SQuAD evaluation script \citep{al2025radqa}.
We additionally credit the best of several predicted spans, which is slightly more lenient to the LLMs, and \citet{soni-etal-2022-radqa} do not state the reference against which each annotator was scored.

\paragraph{Contamination.}
RadQA is distributed only under the PhysioNet Credentialed Health Data License, which prohibits redistribution, but the training data of most evaluated models is undisclosed and we did not test for contamination.
Comparisons between conditions are less affected, because all conditions share the same test items.

\section*{Ethical Considerations}

\paragraph{Data.}
RadQA \citep{soni2022radqaphysionet} is derived from MIMIC-III \citep{PhysioNet-mimiciii-1.4,johnson2016mimic}, which contains de-identified clinical records.
All authors accessed the data through PhysioNet \citep{goldberger2000physiobank} credentialed access, and we used it under the PhysioNet Credentialed Health Data License and its data use agreement.

\paragraph{Responsible use of LLM services.}
We sent MIMIC-derived report text to third-party LLM services only in accordance with PhysioNet's policy on the use of MIMIC data with large language models and online services \citep{physionet2023gpt,physionet2025llm}.
That policy requires zero data retention and advises confirming that services do not use the data for training or subject it to human review; its earlier guidance names Amazon Bedrock and Google Vertex AI among suitable services.
All inference therefore ran through AWS Bedrock and Google Vertex AI with zero data retention confirmed on both platforms: prompts and outputs are neither stored by the provider beyond processing the request, nor used to train or improve its models, nor subject to human review.
No consumer chat interface was used, and no report text was entered into any service outside these configured endpoints.

\paragraph{Data release.}
Our code, including all system prompts and the demonstration-selection procedure, is publicly available at \repourl{}.
It contains no report text: demonstrations are drawn at run time from the credentialed RadQA training split using a fixed seed and question identifiers, so reproducing them requires PhysioNet access.
We do not release report text, model outputs that quote report text, or derived data that would allow reconstruction of the records.

\paragraph{Use of AI assistants.}
We used an AI assistant (Claude Code, Anthropic) to help run the experiments and write the manuscript.
The authors verified all code, results and text, and take full responsibility for the content.

\paragraph{Intended use.}
This work proposes an evaluation method and applies it to 13 LLMs; it does not propose a system for clinical use.
None of the evaluated configurations, including the consensus gate, has been validated for clinical decision-making, and the error rates we report (\far{} up to 0.403 in one condition) show why none should be used to answer questions about patient records without review.
The method is meant for the evaluation that should precede any deployment decision; passing it is not evidence of clinical safety, which would also require local validation and clinical adjudication.

\bibliography{custom}

\appendix

\section{Models and Inference Details}

Table~\ref{tab:models} lists the evaluated models.
Bedrock models were called through the Bedrock Converse API and Gemini models through Vertex AI.

\paragraph{Demonstrations.}
Demonstrations are drawn from the RadQA training split after removing reports longer than 4,000 characters and shuffling with seed 20260719.
The first answerable Findings question (\texttt{879321\_1\_1\_O}), the first answerable Impression question (\texttt{974127\_2\_1\_I}) and the first unanswerable question (\texttt{1061153\_1\_1\_O}) are used, in that order, in every prompt of the corresponding condition.
Each is paired with a fixed rationale written for its type (answerable Findings, answerable Impression, unanswerable), not with any model's output.
Prompts and demonstration rendering are in the released code.

\begin{table*}[t]
  \centering\small
  \setlength{\tabcolsep}{4pt}
  \begin{adjustbox}{max width=\linewidth}
  \begin{tabular}{lllc}
    \toprule
    Model & Provider & Model identifier & Release \\
    \midrule
    Claude Opus 5 \citep{claudeopus5}       & AWS Bedrock & \texttt{us.anthropic.claude-opus-5} & 2026-07-24 \\
    Claude Sonnet 5 \citep{claudesonnet5}   & AWS Bedrock & \texttt{us.anthropic.claude-sonnet-5} & 2026-06-30 \\
    Claude Sonnet 4.5 \citep{claudesonnet4-5} & AWS Bedrock & \texttt{us.anthropic.claude-sonnet-4-5-20250929-v1:0} & 2025-09-29 \\
    Claude Haiku 4.5 \citep{claudehaiku4-5}   & AWS Bedrock & \texttt{us.anthropic.claude-haiku-4-5-20251001-v1:0} & 2025-10-15 \\
    Gemini 3.1 Pro \citep{google2026gemini31pro} & Vertex AI & \texttt{gemini-3.1-pro-preview} & 2026-02-19 \\
    Gemini 3 Flash \citep{google2025gemini3flash} & Vertex AI & \texttt{gemini-3-flash-preview} & 2025-12-17 \\
    LLaMA 4 Maverick \citep{meta2025llama4}  & AWS Bedrock & \texttt{us.meta.llama4-maverick-17b-instruct-v1:0} & 2025-04-05 \\
    LLaMA 4 Scout \citep{meta2025llama4}     & AWS Bedrock & \texttt{us.meta.llama4-scout-17b-instruct-v1:0} & 2025-04-05 \\
    DeepSeek V3.2 \citep{deepseekai2025deepseekv32} & AWS Bedrock & \texttt{deepseek.v3.2} & 2025-12-01 \\
    Mistral Large 3 \citep{mistral2025mistral3} & AWS Bedrock & \texttt{mistral.mistral-large-3-675b-instruct} & 2025-12-02 \\
    Ministral 3 14B \citep{liu2026ministral3} & AWS Bedrock & \texttt{mistral.ministral-3-14b-instruct} & 2025-12-02 \\
    GPT-OSS 120B \citep{openai2025gptoss120bgptoss20bmodel} & AWS Bedrock & \texttt{openai.gpt-oss-120b-1:0} & 2025-08-05 \\
    GPT-OSS 20B \citep{openai2025gptoss120bgptoss20bmodel}  & AWS Bedrock & \texttt{openai.gpt-oss-20b-1:0} & 2025-08-05 \\
    \bottomrule
  \end{tabular}
  \end{adjustbox}
  \caption{Evaluated models. Release is the public announcement date from the developer. Mean input and output tokens per condition are in Table~\ref{tab:tokens}.}
  \label{tab:models}
\end{table*}

\begin{table*}[t]
  \centering\small
  \setlength{\tabcolsep}{3.6pt}
  \begin{adjustbox}{max width=\linewidth}
  \begin{tabular}{l cccc cccc cccc}
    \toprule
    & \multicolumn{4}{c}{F1} & \multicolumn{4}{c}{\far{}} & \multicolumn{4}{c}{\oar{}} \\
    \cmidrule(lr){2-5}\cmidrule(lr){6-9}\cmidrule(lr){10-13}
    Model & \kzero{} & \cF{} & \cFI{} & \cFIU{} & \kzero{} & \cF{} & \cFI{} & \cFIU{} & \kzero{} & \cF{} & \cFI{} & \cFIU{} \\
    \midrule
    Gemini 3 Flash    & 0.748 & 0.735 & 0.738 & 0.716 & 0.123 & 0.156 & 0.143 & 0.143 & 0.100 & 0.080 & 0.100 & 0.130 \\
    Gemini 3.1 Pro    & 0.738 & 0.730 & 0.728 & 0.705 & 0.156 & 0.162 & 0.136 & 0.143 & 0.083 & 0.078 & 0.087 & 0.122 \\
    Claude Sonnet 4.5 & 0.717 & 0.723 & 0.733 & 0.717 & 0.117 & 0.117 & 0.117 & 0.110 & 0.128 & 0.113 & 0.102 & 0.141 \\
    Claude Opus 5     & 0.714 & 0.716 & 0.720 & 0.713 & 0.162 & 0.188 & 0.201 & 0.182 & 0.057 & 0.043 & 0.035 & 0.054 \\
    LLaMA 4 Maverick  & 0.710 & 0.714 & 0.719 & 0.711 & 0.227 & 0.221 & 0.214 & 0.156 & 0.061 & 0.059 & 0.057 & 0.109 \\
    Claude Sonnet 5   & 0.701 & 0.709 & 0.714 & 0.712 & 0.169 & 0.188 & 0.188 & 0.195 & 0.096 & 0.070 & 0.061 & 0.065 \\
    Claude Haiku 4.5  & 0.701 & 0.688 & 0.701 & 0.685 & 0.156 & 0.136 & 0.104 & 0.104 & 0.135 & 0.159 & 0.163 & 0.189 \\
    Mistral Large 3   & 0.686 & 0.710 & 0.710 & 0.706 & 0.169 & 0.143 & 0.169 & 0.143 & 0.133 & 0.117 & 0.111 & 0.137 \\
    GPT-OSS 120B      & 0.681 & 0.677 & 0.666 & 0.677 & 0.130 & 0.123 & 0.110 & 0.117 & 0.150 & 0.163 & 0.180 & 0.178 \\
    DeepSeek V3.2     & 0.676 & 0.680 & 0.718 & 0.688 & 0.260 & 0.208 & 0.175 & 0.156 & 0.130 & 0.120 & 0.093 & 0.152 \\
    LLaMA 4 Scout     & 0.672 & 0.660 & 0.682 & 0.662 & 0.260 & 0.403 & 0.344 & 0.123 & 0.100 & 0.054 & 0.048 & 0.185 \\
    GPT-OSS 20B       & 0.669 & 0.685 & 0.692 & 0.669 & 0.162 & 0.169 & 0.149 & 0.162 & 0.141 & 0.137 & 0.137 & 0.163 \\
    Ministral 3 14B   & 0.588 & 0.644 & 0.655 & 0.652 & 0.182 & 0.162 & 0.169 & 0.130 & 0.202 & 0.204 & 0.193 & 0.224 \\
    \midrule
    Mean              & 0.692 & 0.698 & 0.706 & 0.693 & 0.175 & 0.183 & 0.171 & 0.143 & 0.117 & 0.108 & 0.105 & 0.142 \\
    \bottomrule
  \end{tabular}
  \end{adjustbox}
  \caption{F1, false-answer rate (\far{}) and over-abstention rate (\oar{}) by demonstration condition. Conditions are cumulative: \cF{} adds an answerable Findings demonstration, \cFI{} an answerable Impression demonstration, and \cFIU{} an unanswerable demonstration.}
  \label{tab:conditions}
\end{table*}

\begin{table*}[t]
  \centering\small
  \setlength{\tabcolsep}{4pt}
  \begin{adjustbox}{max width=\linewidth}
  \begin{tabular}{l ccc ccc cc}
    \toprule
    & \multicolumn{3}{c}{Token F1} & \multicolumn{3}{c}{RadGraph F1} & \multicolumn{2}{c}{Counts} \\
    \cmidrule(lr){2-4}\cmidrule(lr){5-7}\cmidrule(lr){8-9}
    Transition & Total $\Delta$ & Extraction & Answerability & Total $\Delta$ & Extraction & Answerability & New abst. & New FA \\
    \midrule
    \kzero{}$\rightarrow$\cF{} & $+$0.0056 & $+$0.0021 & $+$0.0035 & $+$0.0094 & $+$0.0074 & $+$0.0020 & 134 & 71 \\
     & {\scriptsize [$-$0.0007, $+$0.0120]} & {\scriptsize [$-$0.0011, $+$0.0054]} & {\scriptsize [$-$0.0013, $+$0.0083]} & {\scriptsize [$+$0.0028, $+$0.0161]} & {\scriptsize [$+$0.0026, $+$0.0122]} & {\scriptsize [$-$0.0025, $+$0.0065]} & & \\
    \cF{}$\rightarrow$\cFI{} & $+$0.0080 & $+$0.0025 & $+$0.0056 & $+$0.0141 & $+$0.0095 & $+$0.0046 & 96 & 26 \\
     & {\scriptsize [$+$0.0032, $+$0.0132]} & {\scriptsize [$+$0.0001, $+$0.0050]} & {\scriptsize [$+$0.0010, $+$0.0104]} & {\scriptsize [$+$0.0076, $+$0.0215]} & {\scriptsize [$+$0.0048, $+$0.0149]} & {\scriptsize [$+$0.0001, $+$0.0093]} & & \\
    \cFI{}$\rightarrow$\cFIU{} & $-$0.0126 & $+$0.0001 & $-$0.0127 & $-$0.0101 & $-$0.0008 & $-$0.0093 & 272 & 18 \\
     & {\scriptsize [$-$0.0196, $-$0.0058]} & {\scriptsize [$-$0.0018, $+$0.0021]} & {\scriptsize [$-$0.0195, $-$0.0063]} & {\scriptsize [$-$0.0169, $-$0.0032]} & {\scriptsize [$-$0.0034, $+$0.0019]} & {\scriptsize [$-$0.0153, $-$0.0035]} & & \\
    \midrule
    \kzero{}$\rightarrow$\cFI{} & $+$0.0136 & $+$0.0048 & $+$0.0087 & $+$0.0235 & $+$0.0171 & $+$0.0064 & 144 & 62 \\
     & {\scriptsize [$+$0.0072, $+$0.0200]} & {\scriptsize [$+$0.0015, $+$0.0081]} & {\scriptsize [$+$0.0034, $+$0.0144]} & {\scriptsize [$+$0.0149, $+$0.0325]} & {\scriptsize [$+$0.0104, $+$0.0244]} & {\scriptsize [$+$0.0011, $+$0.0120]} & & \\
    \bottomrule
  \end{tabular}
  \end{adjustbox}
  \caption{Decomposition of the change in mean score between conditions (Eq.~\ref{eq:decomp}), pooled over 13 models, with 95\% report-cluster bootstrap CIs (2,000 replicates) below each estimate. Extraction and answerability terms sum to the total. New abst.: answerable questions newly declined; New FA: unanswerable questions newly answered (new false answers). Counts are pooled over models and are the same for both metrics.}
  \label{tab:decomp}
\end{table*}

\begin{table*}[t]
\begin{minipage}[t]{0.46\textwidth}
\centering\small
  \setlength{\tabcolsep}{4pt}
  \begin{adjustbox}{max width=\linewidth}
  \begin{tabular}{lcccc}
    \toprule
    Model & \kzero{} & \cF{} & \cFI{} & \cFIU{} \\
    \midrule
    Gemini 3 Flash & 0.123 & 0.156 & 0.143 & 0.143 \\
    Gemini 3.1 Pro & 0.156 & 0.162 & 0.136 & 0.143 \\
    Claude Sonnet 4.5 & 0.117$^{r}$ & 0.117$^{r}$ & 0.117 & 0.110 \\
    Claude Opus 5 & 0.162 & 0.188$^{*}$ & 0.201$^{*}$ & 0.182$^{*}$ \\
    LLaMA 4 Maverick & 0.227$^{*}$ & 0.221$^{*}$ & 0.214$^{*}$ & 0.156 \\
    Claude Sonnet 5 & 0.169 & 0.188$^{*}$ & 0.188$^{*}$ & 0.195$^{*}$ \\
    Claude Haiku 4.5 & 0.156 & 0.136 & 0.104$^{r}$ & 0.104$^{r}$ \\
    Mistral Large 3 & 0.169 & 0.143 & 0.169$^{*}$ & 0.143 \\
    GPT-OSS 120B & 0.130 & 0.123 & 0.110 & 0.117 \\
    DeepSeek V3.2 & 0.260$^{*}$ & 0.208$^{*}$ & 0.175$^{*}$ & 0.156 \\
    LLaMA 4 Scout & 0.260$^{*}$ & 0.403$^{*}$ & 0.344$^{*}$ & 0.123 \\
    GPT-OSS 20B & 0.162 & 0.169 & 0.149 & 0.162 \\
    Ministral 3 14B & 0.182 & 0.162 & 0.169$^{*}$ & 0.130 \\
    \midrule
    Cochran's $Q$ (df $=12$) & 69.2 & 177.3 & 134.0 & 36.5 \\
    \bottomrule
  \end{tabular}
  \end{adjustbox}
  \caption{\far{} by model and condition. $^{r}$Reference (lowest-\far{}) model; $^{*}$significantly higher than the reference under both tests ($q < 0.05$). All Cochran's $Q$ tests have $p \leq 0.0003$.}
  \label{tab:far-all}
\end{minipage}\hfill
\begin{minipage}[t]{0.50\textwidth}
\centering\small
  \setlength{\tabcolsep}{3pt}
  \begin{adjustbox}{max width=\linewidth}
  \begin{tabular}{lcccc}
    \toprule
    Model & \kzero{} & \cF{} & \cFI{} & \cFIU{} \\
    \midrule
    Gemini 3 Flash & 852 / 552 & 1,145 / 587 & 1,322 / 611 & 1,621 / 654 \\
    Gemini 3.1 Pro & 852 / 361 & 1,145 / 306 & 1,322 / 323 & 1,621 / 334 \\
    Claude Sonnet 4.5 & 1,510 / 159 & 1,844 / 146 & 2,064 / 145 & 2,410 / 144 \\
    Claude Opus 5 & 1,747 / 174 & 2,216 / 168 & 2,538 / 165 & 3,014 / 161 \\
    LLaMA 4 Maverick$^\ddagger$ & 822 / 87 & 1,102 / 77 & 1,270 / 77 & 1,559 / 84 \\
    Claude Sonnet 5 & 1,815 / 170 & 2,284 / 161 & 2,606 / 158 & 3,082 / 159 \\
    Claude Haiku 4.5 & 1,510 / 157 & 1,844 / 142 & 2,064 / 142 & 2,410 / 141 \\
    Mistral Large 3 & 1,201 / 268 & 1,509 / 173 & 1,662 / 130 & 1,959 / 123 \\
    GPT-OSS 120B & 1,184 / 242 & 1,473 / 245 & 1,643 / 246 & 1,937 / 243 \\
    DeepSeek V3.2 & 967 / 211 & 1,253 / 180 & 1,417 / 184 & 1,709 / 183 \\
    LLaMA 4 Scout$^\ddagger$ & 822 / 89 & 836 / 76 & 1,004 / 75 & 1,293 / 71 \\
    GPT-OSS 20B$^\S$ & 1,184 / 251 & 1,549 / 295 & 1,832 / 319 & 2,271 / 372 \\
    Ministral 3 14B & 1,201 / 105 & 1,494 / 96 & 1,662 / 94 & 1,959 / 94 \\
    \midrule
    \emph{Model mean} & 1,205 / 217 & 1,515 / 204 & 1,724 / 205 & 2,065 / 212 \\
    \bottomrule
  \end{tabular}
  \end{adjustbox}
  \caption{Token usage by model and condition. Mean input / output tokens per row (all 894 rows of the test file; \S\ref{sec:data}), as reported by each provider's API. Counts use each model's own tokenizer, so they are comparable across conditions within a model but not across model families. $^\ddagger$Bedrock-reported input tokens for the LLaMA~4 models are inconsistent: Scout reports about 266 fewer tokens than Maverick in every demonstration condition, and re-querying both with an identical prompt gave equal, lower counts. A probe confirmed that Scout received all demonstrations; treat LLaMA~4 input-token counts as approximate. $^\S$Token means are inflated by retried calls (counts exactly 2--3$\times$ those of GPT-OSS 120B on 29/74/109 of 894 rows at \cF{}/\cFI{}/\cFIU{}); answers are unaffected.}
  \label{tab:tokens}
\end{minipage}
\end{table*}

\begin{table*}[t]
\begin{minipage}[t]{0.48\textwidth}
\centering\small
  \setlength{\tabcolsep}{3pt}
  \begin{adjustbox}{max width=\linewidth}
  \begin{tabular}{lcccc}
    \toprule
    Measure & \kzero{} & \cF{} & \cFI{} & \cFIU{} \\
    \midrule
    SQuAD F1 & $-$0.59 {\scriptsize (.036)} & $-$0.23 {\scriptsize (.46)} & $-$0.04 {\scriptsize (.89)} & $+$0.25 {\scriptsize (.41)} \\
    Answerable F1 & $-$0.28 {\scriptsize (.35)} & $+$0.35 {\scriptsize (.24)} & $+$0.48 {\scriptsize (.10)} & $+$0.55 {\scriptsize (.053)} \\
    Extraction F1 & $-$0.47 {\scriptsize (.105)} & $-$0.59 {\scriptsize (.033)} & $-$0.49 {\scriptsize (.092)} & $-$0.43 {\scriptsize (.14)} \\
    \oar{} & $-$0.08 {\scriptsize (.78)} & $-$0.58 {\scriptsize (.037)} & $-$0.72 {\scriptsize (.005)} & $-$0.69 {\scriptsize (.009)} \\
    \bottomrule
  \end{tabular}
  \end{adjustbox}
  \caption{Spearman correlation with \far{} across the 13 models ($p$ in parentheses). Answerable F1 scores abstentions on answerable questions as 0; extraction F1 averages over answerable questions the model answered (\S\ref{sec:scoring}).}
  \label{tab:corr}

\vspace{\floatsep}
\centering\small
  \setlength{\tabcolsep}{4pt}
  \begin{adjustbox}{max width=\linewidth}
  \begin{tabular}{lccc}
    \toprule
    Condition & Extraction & Propensity & Difficulty \\
    \midrule
    \kzero{} & 1.04 {\scriptsize [0.92, 1.18]} & 1.21 {\scriptsize [1.01, 1.45]} & 0.84 {\scriptsize [0.57, 1.24]} \\
    \cF{}    & 0.94 {\scriptsize [0.84, 1.04]} & 1.23 {\scriptsize [0.94, 1.60]} & 0.88 {\scriptsize [0.55, 1.39]} \\
    \cFI{}   & 0.99 {\scriptsize [0.86, 1.15]} & 1.31 {\scriptsize [1.03, 1.65]} & 0.83 {\scriptsize [0.56, 1.22]} \\
    \cFIU{}  & 0.99 {\scriptsize [0.89, 1.09]} & 1.10 {\scriptsize [0.79, 1.53]} & 0.91 {\scriptsize [0.59, 1.42]} \\
    \bottomrule
  \end{tabular}
  \end{adjustbox}
  \caption{Within-model predictors of a false answer: odds ratio per SD [95\% CI] from a logistic GEE with model fixed effects and report clustering (1,596--1,616 model $\times$ unanswerable-question instances per condition, 52 reports). Extraction: the model's mean F1 on the report's answerable questions it answered; Propensity: the fraction of those questions it answered; Difficulty: the other 12 models' mean F1 on the report (\S\ref{sec:scoring}).}
  \label{tab:gee}
\end{minipage}\hfill
\begin{minipage}[t]{0.48\textwidth}
\centering\small
  \setlength{\tabcolsep}{4pt}
  \begin{adjustbox}{max width=\linewidth}
  \begin{tabular}{lccc}
    \toprule
    Condition, rule & $\Delta$F1 & $\Delta$\far{} & $\Delta$\oar{} \\
    \midrule
    \kzero{}, own-$t$ & $+$0.005 & $-$0.040 & $+$0.010 \\
     & {\scriptsize [$-$0.000, $+$0.009]} & {\scriptsize [$-$0.057, $-$0.026]} & {\scriptsize [$+$0.004, $+$0.018]} \\
    \kzero{}, pooled-$t$ & $+$0.004 & $-$0.042 & $+$0.012 \\
     & {\scriptsize [$-$0.003, $+$0.009]} & {\scriptsize [$-$0.063, $-$0.016]} & {\scriptsize [$+$0.002, $+$0.027]} \\
    \cF{}, own-$t$ & $+$0.005 & $-$0.032 & $+$0.006 \\
     & {\scriptsize [$-$0.001, $+$0.008]} & {\scriptsize [$-$0.044, $-$0.023]} & {\scriptsize [$+$0.001, $+$0.015]} \\
    \cFI{}, own-$t$ & $+$0.003 & $-$0.031 & $+$0.009 \\
     & {\scriptsize [$-$0.003, $+$0.006]} & {\scriptsize [$-$0.045, $-$0.018]} & {\scriptsize [$+$0.003, $+$0.023]} \\
    \cFIU{}, own-$t$ & $-$0.000 & $-$0.011 & $+$0.005 \\
     & {\scriptsize [$-$0.004, $+$0.001]} & {\scriptsize [$-$0.018, $-$0.004]} & {\scriptsize [$+$0.002, $+$0.012]} \\
    \bottomrule
  \end{tabular}
  \end{adjustbox}
  \caption{Held-out effect of the cross-fitted consensus gate: model-mean change vs.\ the ungated model, with 2.5--97.5\% intervals over 200 report splits below each estimate. Pooled-$t$ uses one threshold for all models (modal $t = 4$).}
  \label{tab:gate}
\end{minipage}
\end{table*}

\section{Statistical Details}
\label{app:stats}

\paragraph{Condition effects and intervals.}
Model-level CIs use a report-level cluster bootstrap (10,000 replicates; 2,000 for decomposition terms).
Per-model condition effects use a Wilcoxon signed-rank test on the 104 paired report-mean scores, with BH correction across the 13 models within each transition; per-model extraction terms are tested in the same way on questions in $\mathcal{E}$.
Over model means, condition effects are tested with a Wilcoxon test across the 13 models.

\paragraph{False-answer rate.}
Cochran's Q tests whether the 13 models differ in \far{} within a condition.
Each model is compared with the lowest-\far{} model on the same 154 unanswerable questions by an exact McNemar test and a report-level sign-flip permutation test (10,000 permutations), BH-corrected within condition (Table~\ref{tab:far-all}).
An unclustered McNemar screen of per-question correctness across all 78 model pairs (BH across 312 tests) finds 32 pairs that differ on F1 $\geq$ 0.5 at \kzero{}; Gemini~3 Flash is not separable from Gemini~3.1 Pro, Claude Opus~5, Claude Sonnet~4.5 or LLaMA~4 Maverick ($q = 0.136$--$1.000$).

\paragraph{Item-level models.}
Selectivity regresses prior score on a newly-abstained indicator (Gaussian GEE, exchangeable within-report correlation); pooled fits include model fixed effects and are checked by leaving out one model at a time, and per-model fits (models with at least 5 new abstentions) are BH-corrected across models.
Within-model predictors of false answers (\S\ref{sec:scoring}) use a logistic GEE over model $\times$ unanswerable-question instances (1,596--1,616 per condition), with extraction quality and answering propensity centred within model, model fixed effects and the report-difficulty control.

\paragraph{Correlations across models.}
Spearman $\rho$ with a 10,000-replicate bootstrap over models and leave-one-model-out $p$-values (Table~\ref{tab:corr}).

\paragraph{Error-taxonomy thresholds.}
The no-overlap rate at \kzero{} is 1.64\%, 1.88\% and 3.41\% for F1 thresholds of 0.05, 0.10 (used) and 0.20; its decline from \kzero{} to \cFI{} holds at every threshold, and the EM-mismatch threshold (0.4--0.6) does not affect it.

\section{Additional Results}

Table~\ref{tab:conditions} gives per-model F1, \far{} and \oar{} in every condition; Table~\ref{tab:decomp} the exact values behind Figure~\ref{fig:decomp}; Table~\ref{tab:far-all} the \far{} tests in every condition; Tables~\ref{tab:corr} and~\ref{tab:gee} the cross-model and within-model analyses of \S\ref{sec:res-far-accuracy}; Table~\ref{tab:gate} the consensus gate; and Table~\ref{tab:tokens} token usage.

\end{document}